# Evaluation and Ranking of Machine Translated Output in Hindi Language using Precision and Recall Oriented Metrics

Aditi Kalyani[1], Hemant Kumud[2], Shashi Pal Singh[3], Ajai Kumar[4], Hemant Darbari[5]

## Abstract

*Evaluation plays a crucial role in development of Machine translation systems. In order to judge the quality of an existing MT system i.e. if the translated output is of human translation quality or not, various automatic metrics exist. We here present the implementation results of different metrics when used on Hindi language along with their comparisons, illustrating how effective are these metrics on languages like Hindi (free word order language).*

## 1. Introduction

Evaluation of Machine Translation (MT) has historically proven to be a very difficult exercise. The difficulty stems primarily from the fact that translation is more of an art than science; majority of the sentences can be translated in many adequate ways. Consequently, there is no *golden standard a*gainst which a translation can be assessed.

MT Evaluation strategies were initially proposed by Miller and Beeber-center in 1956 followed by Pfaffine in 1965. At the start MT evaluation was performed only by human judges. This process, however, was time-consuming and highly prejudiced. Hence arose the requirement for automation i.e., for fast, objective, and reusable methods of evaluation, the results of which are not biased or subjective at all. To this end, several metrics for automatic evaluation have been proposed and have been accepted actively.

**Aditi Kalyani**, Department of Computer Science, Banasthali Vidyapith, Banasthali, India.
**Hemant Kumud,** Department of Computer Science, Banasthali Vidyapith, Banasthali, India.
**Shashi Pal Singh**, AAI, CDAC, Pune, India
**Ajai Kumar**, AAI, CDAC, Pune, India.
**Dr. Hemant Darbari**, ED, CDAC, Pune, India.

Automatic MT evaluation started with introduction of BLEU proposed by Paninani et al in 2001. Following IBM's metric (BLEU), DARPA designed NIST in 2002, Lavie and Denkowski proposed METEOR in 2005.

In this paper we discuss the implementation results of various metrics when used with Hindi language along with their comparisons, depicting their effectiveness on languages like Hindi (free word order language).
In section 2 we briefly provide the amalgamated study of human and automatic evaluation strategies also giving a brief review of the work done in the area. Section 3 describes effectiveness of BLEU metric for Hindi and other morphologically rich languages. Section 4 presents the issues in evaluation of free word order languages. Section 5 compares the performance of METEOR with METEOR-HINDI specifically tailored for Hindi. Section 6 shows comparative results of section 3 and 5. Section 7 concludes the work done along with future trends.

## 2. Human Vs. Automatic Evaluation

Evaluation is usually done in two ways: Human and automatic.

### 2.1 Human Evaluation
Evaluation of machine translated output by human is a fusion of values: fluency, adequacy and fidelity (Hovy, 1999; White and O'Connell, 1994). Adequacy deals with the meaning of translated output i.e. if both candidate and reference mean the same thing or not, fluency involves both the language rules correctness and phrase word choice and fidelity is the amount of information retained in translated output in comparison to candidate.

**Table 1: Human Criterion for Rating**

| Rating | Translation-Quality |
|---|---|
| 5 | Excellent |
| 4 | Good |
| 3 | Understandable |
| 2 | Barely Understandable |
| 1 | Unacceptable |

Table 1 gives possible evaluating criteria by humans to measure the score for a translation.

In human evaluation there are two types of evaluators: Bilingual, those who understand both source and target languages and others are monolingual i.e. understanding only target language. Here, the human evaluator looks at the translation and judges it to check that if it is correct or not. One of the most important peculiarities of human evaluation is that two human evaluators when judging the same text could give two different evaluations, as might the same evaluator at different moments (even for exact matches).Which means that human criteria for evaluation of Machine output is subjective. Also human evaluations are non reusable, expensive and time consuming. To overcome these situations we need an automatic system which can perform faster and give the output if not same but comparable to human output and can be reused over and over.

### 2.2 Automatic Evaluation

Evaluations of machine translation by human are extensive but very expensive and time consuming. Human evaluations can take months and the worst part is they can not be reused. A good evaluation metric in general should, 1. Be Quick 2. Be Inexpensive 3. Be language-independent 4. Correlate highly with human evaluation 5. Have little marginal cost per run [1] Apart from above mentioned characteristics a metric should be consistent (Same MT system on similar texts should produce analogous scores), general (appropriate to different MT tasks in a wide range of fields and situations) and reliable (MT systems that score alike can be trusted to perform likewise).

In general all automatic metrics are based one of the following to calculate scores [14]:
- Number of changes required to make candidate as reference in terms of number of insertions, deletions and substitutions are counted i.e. **Edit Distance**
- Total number of matched unigrams are divided by the total length of candidate i.e. **Precision**
- Total number of matched unigrams are divided by the total length of reference i.e. **Recall**
- Both precision and recall scores are used collectively i.e. **F-measure**

Various metrics for automatic evaluation have been proposed and the research is never ending. Some of the metrics are as described in Table 2.

**Table 2: Automatic Evaluation Metrics**

| METRIC | FEATURE |
|---|---|
| BLEU (Papineni et al., 2001)[1] | Based on average of matching n-grams between candidate and reference |
| NIST (Doddington, 2002)[3] | Calculate matched n-grams of sentences and attach different weights to them |
| GTM (Turian et al., 2003)[14] | Computes precision recall and f-measure in terms of maximum unigram matches. |
| ROUGE (Lin and Hovy, 2003)[14] | Creates the summary & compares it with the summary created by human. (Recall oriented) |
| METEOR (Banerjee & Lavie, 2005)[4] [10] {latest modification: 2012} | Based on various modules (Exact Match, Stem Match, Synonym Match and POS Tagger) |
| BLANC (Lita et al., 2005)[9] | Based on features of BLEU and ROUGE |
| TER (Snover et al., 2006)[14] | Metric for measuring mismatches |
| ROSE (Song and Cohn, 2011)[9] | Uses syntactic resemblance (Here Part of Speech) |
| AMBER (Chen and Kuhn, 2011)[9] | Based on BLEU but adds recall, extra penalties, and some text processing variants |
| LEPOR (Han et al., 2012)[8] | Combines sentence length penalty and n-gram position difference penalty. Also uses precision and recall |
| PORT (Chen et al., 2012)[9] | Based on precision, recall, strict brevity penalty, strict redundancy penalty and an ordering measure. |
| METEOR Hindi (Ankush Gupta et al., 2010)[2] | A modified version of the METEOR containing features specific to Hindi |

## 3. BLEU Deconstructed for Hindi

BLEU (*Bilingual Evaluation Understudy*) is n-gram based metric. Here for each n, where n usually ranges from 1 to a maximum of 4, count the number of occurrences of n-grams in the test translation that have a match in the corresponding reference translations. BLEU uses modified n-gram precision in which a reference translation is considered exhausted after a matching candidate word is found.

This is done so that one word of candidate matches to only one word of reference translation. A brevity penalty is introduced to compensate for the possibility of proposing high precision hypothesis translations which are too short. The final BLEU formula [1] is:

$$S_{bleu} = e^{\left(1-\frac{r}{t}\right)} e^{\sum_{i=1}^{N} w_i \log(S_i)}$$

Geometric averaging on n-gram scores zero if any of the n-gram is zero. Since the precision of 4-gram is many times 0, the BLEU score is generally computed over the test corpus rather than on the sentence level. Many enhancements have been done on the basic BLEU algorithm, e.g. Smoothed BLEU (Lin and Och 2004) etc. to provide better results.

Figure 1 shows the implementation details of BLEU. It is a flowchart of entire procedure, candidate and reference are picked from database and then words from a candidate translation that match with a word in the reference translation (human translation) are counted, and then divided by the number of words in the candidate translation ($S_i$).

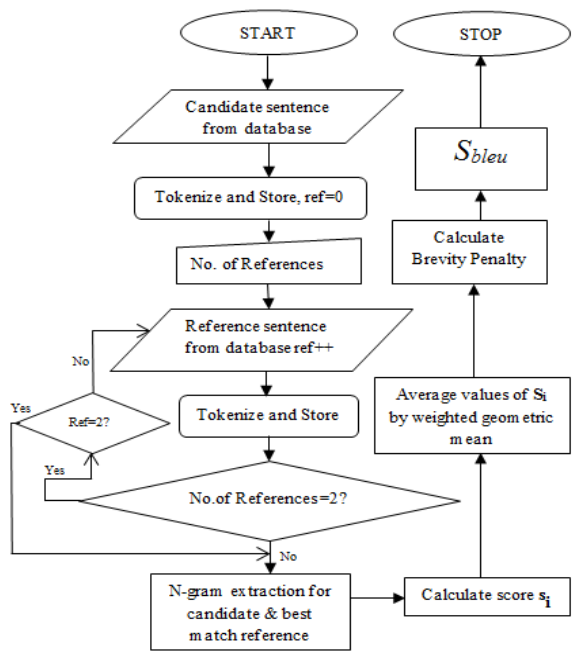

Figure 1: BLEU Flowchart

### 3.1 N-gram Comparison with BLEU Score
The BLEU score ranges from 0 to 1. The more the number of matching n-grams the higher the BLEU score is. It can be seen from Table 3 that even though sentence is grammatically correct, the BLEU score decreases on increasing degree of n-grams. The scores given in table are based on entire corpus and not on individual sentences; this is because the scores over entire corpus are much more reliable and accurate.

Table 3: N-gram Comparison

| N-Grams | Total | Matched | BLEU Score |
|---|---|---|---|
| Bi-gram | 2549 | 1172 | 0.46 |
| Tri-gram | 2548 | 790 | 0.31 |
| 4-gram | 2547 | 356 | 0.14 |

BLEU score can be different for translations that are semantically close but have change in the order of arrangement of words. One such example is given below:

C1: सीता ने अलमारी में रखा हुआ कटा सेब खाया|

R1: अलमारी में रखा कटा हुआ सेब सीता ने खाया|

R2: अलमारी में रखा हुआ कटा सेब सीता ने खाया|

R4: रखा हुआ कटा सेब अलमारी में सीता ने खाया|

The above translations have same meaning and use same words only with different permutations to form sentence. The BLEU score for varying n-gram arrangements of above sentences is pictured in Figure 1. It can be seen that as the value of n is increased the BLEU score decreases.

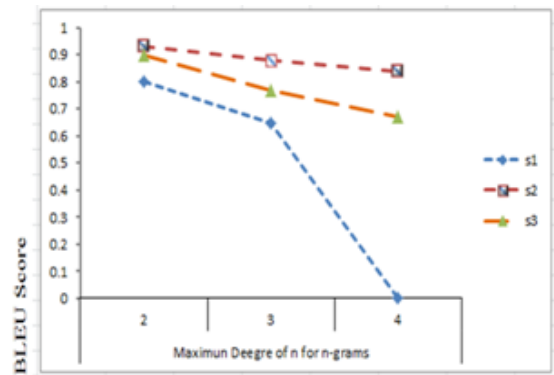

Figure 2: N-gram Graph

For all the values of n, the performance of BLEU for language like Hindi is best when n=2 because there is no guarantee of word sequences in free word order languages. Hence BLEU doesn't really go well for Hindi without making some severe changes to the metric.

## 4. Why not BLEU for Hindi

There are various characteristic Features of Hindi like morphological richness, no ordering of words, etc. which makes it hard to handle when performing any kind of MT related operations. If BLEU is used for free word order, morphologically rich language (the way words are constructed with stems, prefixes and suffixes) and language laden with synonyms, its quality may get down due to only exact word match unigrams and the score will not be of much significance. Also many key features of the language will not be taken into account leading to incorrect evaluation and sometimes giving useless results. Hence suitability of BLEU for the languages like Hindi has been a big issue since the very beginning.

## 5. METEOR Vs. METEOR-Hindi

METEOR was designed to address the weaknesses in BLEU. It is based on a word-to-word alignment between the machine-generated translation and the reference translation. Every unigram in the test translation should map to zero or one unigram in the reference sentence. If there are two alignments with the same number of mapping, the alignment is chosen with less number of intersections of the two mappings. The score is equal to the harmonic mean of unigram precision and unigram recall. It also has several features that are not found in other metrics such as stemming, synonymy matching etc.
Original METEOR consists of:
1) Exact Match mapping words that are exactly same;
2) Stem Match links words that share the same stem;
3) Synonym Match mapping unigrams that are synonyms of each other.

METEOR-Hindi includes following additional modules to make more efficient for Hindi:
1) The local word group (LWG) consisting of a content word and its associated function words;
2) Clause Match- Clause is defined as a phrase containing at least a verb and a subject;
3) POS matcher computes the number of matching words with same POS tag.
METEOR also includes a fragmentation penalty that accounts for how well-ordered the matched unigrams of the machine translation are with respect to the reference. The alignment between machine translation and reference translation is obtained through mapping modules that apply sequentially, linking unigrams that have not been mapped by the previous modules. [2]

## 6. Comparative Results

Table 4 shows the performance of various metrics on a corpus of 1000 sentences. Here the score for BLEU is least and the best performance is obtained by METEOR-Hindi which correlates well with human judgment. Hence Meteor –Hindi outperforms others in most of the cases but still there is scope for improvements because of the Morphological richness of Hindi language. The value for human score is obtained by using the criteria mentioned in table 1.

**Table 4: Comparative scores**

| Metric | Scores |
|---|---|
| BLEU | 0.14 |
| METEOR | 0.34 |
| METEOR-Hindi | 0.44 |
| Human | 0.69 |

Figure 3 shows the scores of various translation engines for BLEU, METEOR and METEOR-Hindi. Statistics were calculated on data set of 500 sentences. METEOR-Hindi usually performs better than other two but in some cases the scores are not that good even when compared with original BLEU.

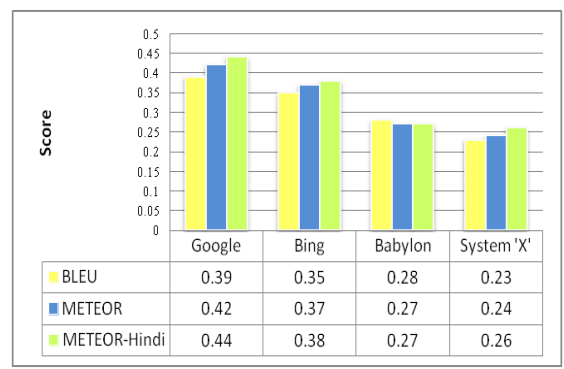

**Figure 3: Scores for different Translators**

## 7. Conclusions

The results of BLEU for Hindi do not correlate that well with scores given by humans. BLEU does not work properly for free word order languages. The result for 2-grams is highly impractical because it does not take into account fluency, which is main component of evaluation. The best results for Hindi using BLEU are of 3-grams due to balance between adequacy and fluency.

METEOR original is good but in METEOR-Hindi after addition of more modules the accuracy is quite better. Several metrics exist apart from these but the performance of METEOR-Hindi beats all for Hindi language.

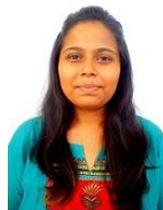
**Ms. Aditi Kalyani** is a Research Scholar and doing her internship from C-DAC Pune under M.Tech, Computer Science II year Curriculum pursued by Banasthali University, Rajasthan, India. She has completed her B.E degree in Computer Science from University of Pune, Maharashtra.

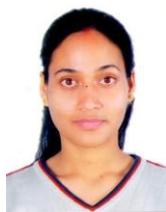 **Mrs Hemant Kumud** is a Research Scholar and doing her internship from C-DAC, Pune under M.Tech, Computer Science II year Curriculum which is being pursued from Banasthali University, Rajasthan, India. She has completed her Master of Computer Application (MCA) from Maharshi Dayanand Univ., Rohtak.

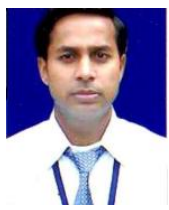 **Mr. Shashi Pal Singh** is working as STO, AAI Group, C-DAC, Pune. He has completed his B.Tech and M.Tech in Computer Science & Engg. and has published various national & international papers. He is specialised in Natural Language Processing (NLP), Machine assisted Translation (MT), Cloud Computing and  Mobile Computing.

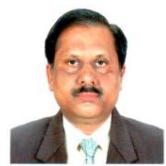 **Mr. Ajai Kumar** is working as Associate Director and Head, AAI Group, C-DAC, Pune. He is handling various projects in the area of Natural Language Processing, Information Extraction and Retrieval, Intelligent Language Teaching/Tutoring, Speech Technology [Synthesis & Recognition ASR], Mobile Computing, Decision Support Systems & Simulations and has published various national & international papers.

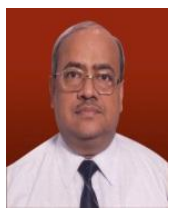 **Dr. Hemant Darbari** is working as Executive Director in C-DAC, Pune. He is one of the founding members of C-DAC, an R&D institute set up by the Department of Electronics and Information Technology; Govt. of India for carrying out advanced research in new and emerging technological domains. He has to his credit, 85 Technical Papers that have been published in national & international Journals & Conference Proceedings.